\documentclass[lettersize,journal]{IEEEtran}
\usepackage{amsmath,amsfonts}
\usepackage{array}
\usepackage[caption=false,font=normalsize,labelfont=sf,textfont=sf]{subfig}
\usepackage{textcomp}
\usepackage{stfloats}
\usepackage{url}
\usepackage{verbatim}
\usepackage{graphicx}
\def\BibTeX{{\rm B\kern-.05em{\sc i\kern-.025em b}\kern-.08em
    T\kern-.1667em\lower.7ex\hbox{E}\kern-.125emX}}
\usepackage{balance}

\usepackage{caption}
\usepackage{algorithm}
\usepackage{algorithmic}
\usepackage{graphicx}
\usepackage{enumerate}
\usepackage{amsthm}
\usepackage{epstopdf}
\usepackage{amsmath,amssymb,epsfig,graphicx,slidesec,colortbl}
\usepackage[absolute,overlay]{textpos}
\usepackage{multimedia,algorithm,multirow,url}
\usepackage{cite}
\floatname{algorithm}{Algorithm 1}
\usepackage{bbm}
\usepackage{subfig}
\usepackage{color}
\usepackage{url}
\usepackage{enumitem}
\usepackage{balance}
\usepackage{tcolorbox}

\usepackage[font={small}]{caption}

\begin{document}
\title{Cross-Silo Federated Learning:\\ Challenges and Opportunities}

\author{Chao Huang,  Jianwei Huang,\IEEEmembership{~Fellow,~IEEE}, and Xin Liu,\IEEEmembership{~Fellow,~IEEE}}

\markboth{Submitted to IEEE Communications Magazine for Possible Publication}%
{Shell \MakeLowercase{\textit{et al.}}: A Sample Article Using IEEEtran.cls for IEEE Journals}


\maketitle

\begin{abstract}
	Federated learning (FL) is an emerging technology that enables the training of machine learning models from multiple clients while keeping the data distributed and private. Based on the participating clients and the model training scale, federated learning can be classified into two types: cross-device FL where clients are typically mobile devices and the client number can reach up to a scale of millions; cross-silo FL where clients are organizations or companies and the client number is usually small (e.g., within a hundred). While existing studies mainly focus on cross-device FL, this paper aims to provide an overview of the cross-silo FL. More specifically, we first discuss applications of cross-silo FL and outline its major challenges. We then provide a systematic overview of the existing approaches to the challenges in cross-silo FL by focusing on their connections and differences to cross-device FL. Finally, we discuss future directions and open issues that merit research efforts from the community.
\end{abstract}


\section{Introduction}
Federated learning (FL) is a distributed machine learning
scheme where a bunch of clients collaboratively train
a global model under the coordination of a central server. Clients train models using their private local data and they
only need to upload model updates (e.g., represented by parameters
or gradients) to the server. Based on the participating clients and training scale, FL can be divided into two types: cross-device FL and cross-silo FL. In
cross-device FL, clients are small distributed entities
(e.g., smartphones, wearables, and edge devices), and each client is likely to have a relatively small amount of local data. Hence, for
cross-device FL to succeed, it usually requires a large number (e.g., up to millions) of edge devices to
participate in the training process. In cross-silo FL, however, clients are typically companies or organizations (e.g., hospitals and
banks). The number of participants is small (e.g., from two to a hundred), and each client is expected to participate in the entire training process. 

Previous studies focus on cross-device FL.  Interested readers can refer to  \cite{yang2019federated,kairouz2021advances} for excellent surveys.
The focus of this paper, however, is cross-silo FL. Practical examples of cross-silo FL abound \cite{li2020review}. In the health care domain,  different medical institutes collaborate to train disease prediction models. For example, Owkin collaborates with pharmaceutical companies to train models for drug discovery based
on sensitive screening datasets. In the finance realm, financial organizations cooperate to train prediction models and provide customized services. For instance, WeBank and Swiss Re collectively perform data analysis and provide
financial and insurance services. In the transportation sector, companies who own distributed traffic data train a global model to predict future traffic flow.



 While there has an emerging stream of research on cross-silo FL, a systematic overview is missing in the current literature. We aim to fill this gap by providing a first overview of cross-silo FL. In particular, we overview the key challenges and solutions in cross-silo FL by explaining its connections to and focusing on its differences from cross-device FL. 

The remainder of this paper is as follows. In Section \ref{Overview}, we introduce the problem formulation and taxonomy. From Section \ref{effectiveness_efficiency} to Section \ref{Cooperation_incentive}. we discuss the key challenges and existing approaches in cross-silo FL while elaborating on the key differences to cross-device FL. We discuss open issues and future directions in Section \ref{future_work}.

\begin{figure}[!t]
	\centering
	\includegraphics[width=3.5in]{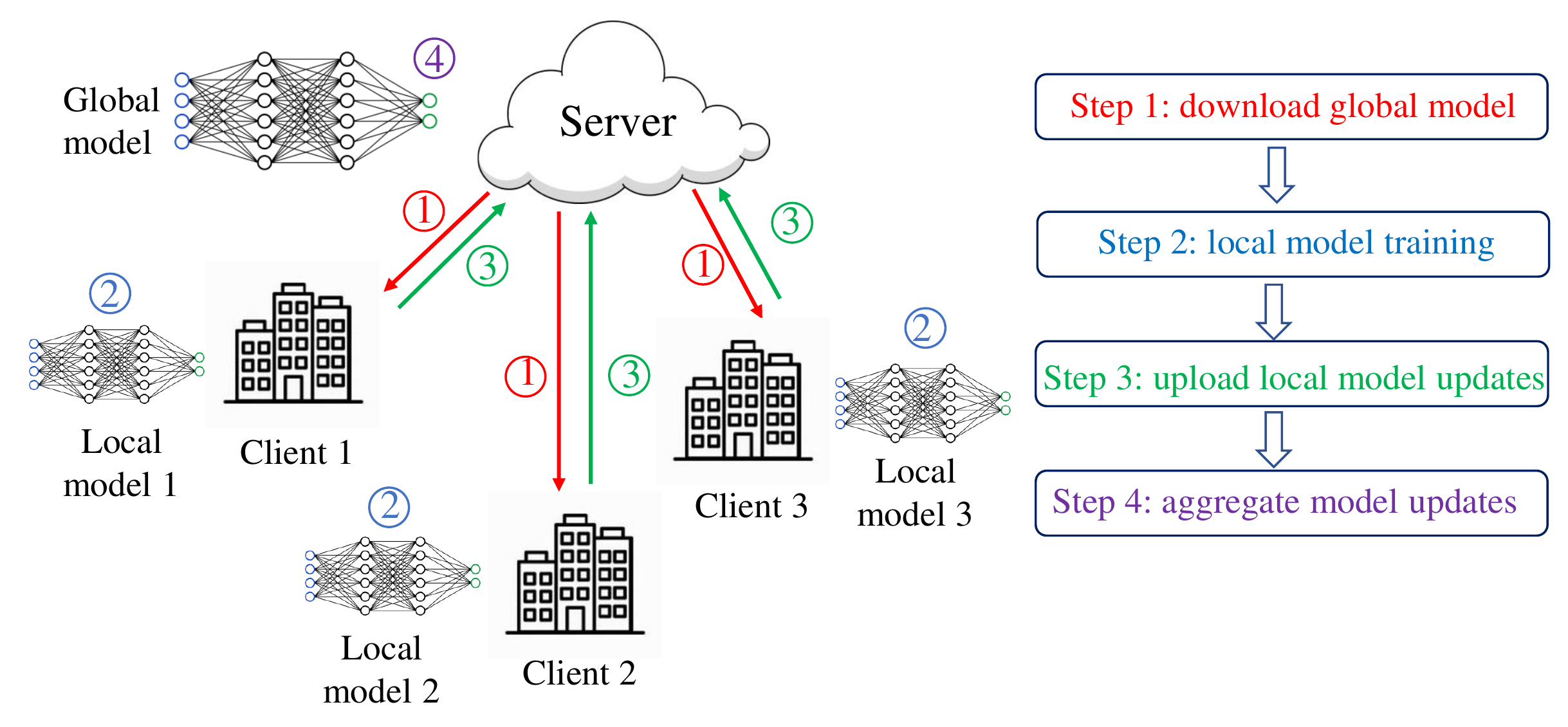}
	\caption{A typical cross-silo FL process.}
	\label{FL}
\end{figure} 
\section{Problem Definition and Taxonomy}\label{Overview}
In this section, we first give a problem definition for cross-silo FL in subsection \ref{formulation}, and then provide a taxonomy for the key challenges and solutions in subsection \ref{challenges_solutions}.
\subsection{Problem Definition}\label{formulation}
A typical cross-silo FL process consists of multiple rounds. In each round, there are four steps (see also Fig. \ref{FL}):
\begin{itemize}
	\item \textbf{Step 1}: The clients download the global model generated from the previous round from the central server (in round one, the downloaded model is randomly initialized).
	\item \textbf{Step 2}: The clients train the downloaded models using their private local data sets and derive updated local models.
	\item \textbf{Step 3}: The clients upload their local model updates to the central server.
	\item \textbf{Step 4}: The server aggregates the uploaded model updates and generates a new global model to be sent to the clients in the next round.
\end{itemize}

The cross-silo FL terminates when the global model converges or the number of training rounds exceeds a predefined threshold.

\subsection{Challenge Overview}\label{challenges_solutions}
\begin{figure*}
	\center
	\includegraphics[width=0.95\textwidth]{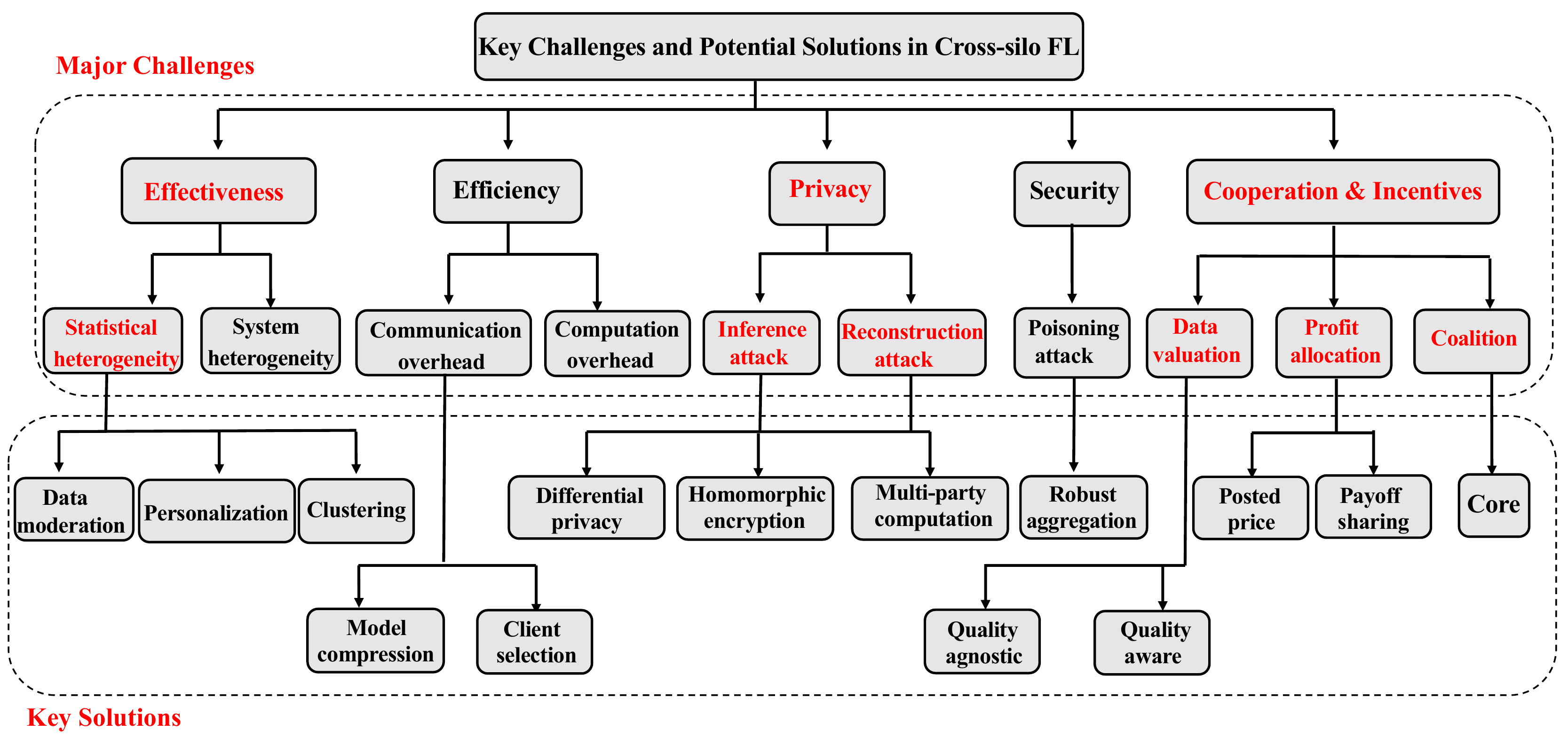}
	\caption{A taxonomy of challenges and solutions in cross-silo FL.}
	\label{taxonomy}
\end{figure*}

There are three aspects of challenges regarding the practical design of cross-silo FL as follows:
\begin{itemize}
	\item \textbf{Effectiveness and Efficiency}: One major challenge is how to execute cross-silo FL effectively and efficiently.  Effectiveness refers to obtaining satisfactory global (and local) models accounting for various client heterogeneity, and efficiency refers to obtaining an effective model fast with a low cost. 
	\item \textbf{Privacy and Security}: Another important challenge is privacy and security. Privacy is concerned with protecting clients' local data from being leaked, while security pertains to detecting and refraining adversaries from jeopardizing the model training process and the model performance. 
	\item \textbf{Cooperation and Incentives}: Unlike edge devices in cross-device FL, organizations or companies in cross-silo FL usually have clear long-term strategic focuses and development goals. This makes the long-term cooperation relationship more possible, as long as we can design a proper incentive mechanism. 
\end{itemize}
Fig. \ref{taxonomy} summarizes the taxonomy of challenges in cross-silo FL, where we further outline the corresponding key solutions (to be detailed in the following sections). In particular, the challenges in red color are the ones that are more critical in cross-silo settings and hence deserve more research attention. 

\section{Effectiveness and Efficiency}\label{effectiveness_efficiency}
The success of cross-silo FL depends on two critical factors: \textit{effectiveness} and \textit{efficiency}. Effectiveness is concerned with training a global model with a desirable performance considering the high heterogeneity of participating clients. Efficiency pertains to training the model with few computation and communication overheads. In this section, we discuss methods to achieve better effectiveness and efficiency. 

\subsection{Optimization of Effectiveness}
Two components crucially affect the effectiveness of FL, i.e., statistical heterogeneity and system heterogeneity. 

\subsubsection{Statistical heterogeneity} This often refers to the non-i.i.d. data of clients \cite{wei2022vertical}. For example, different hospitals may hold non-i.i.d. disease data due to having geographically and demographically varying patients. Different banks can have heterogeneous data since customers from different regions/countries can have diverse economic backgrounds and investment habits.

To tackle the statistical heterogeneity, existing studies have developed three main approaches discussed as follows \cite{zhu2021federated}:
\begin{itemize}
	\item \textbf{Data moderation approaches}: Intuitively, the FL performance degradation with non-i.i.d. data stems from the heterogeneous data distributions. Data moderation aims to tackle this issue by modifying the distributions. One possible method is \textit{data sharing}, where clients' local models are trained by not only the local data but also the shared global data at the server. Another method is \textit{data augmentation}, which replenishes the local data with augmentations based on information from other clients' data distributions. 
	\item \textbf{Personalization approaches}: The global model computed by the server may negatively impact the clients' local model performances in terms of performance and generalization, due to the drift of heterogeneous data. Personalization approaches aim to address this issue by adjusting the local models based on the local tasks. There are several types of personalization approaches. One is \textit{local fine-tuning}, which fine-tunes local models using local data after receiving the global model. Another is \textit{personalized layer}, in which clients' local models have base layers and personalized layers, and only the base layers need to be uploaded for aggregation. The third one is \textit{knowledge distillation}, which aims to transfer knowledge from the server (or other clients) to a specified client to enhance its local model performance on heterogeneous data. 
	\item \textbf{Clustering approaches}: While most FL frameworks assume that there is only one global model, it can be insufficient to learn clients' information when data are highly heterogeneous. Clustering approaches try to mitigate this issue by grouping the clients into different clusters, in which each cluster maintains a server-client framework. In particular, similar clients are allocated to the same cluster, where similarity is measured by the loss values or their uploaded model weights \cite{ouyang2021clusterfl}. 
	\end{itemize}

\subsubsection{System Heterogeneity} Different clients often vary in system characteristics such as hardware, network connectivity, and energy constraints. This can lead to the straggler effect, mostly in cross-device FL. Interested readers can refer to \cite{li2020federated} for a more detailed discussion on tackling system heterogeneity.

\textit{Remark}: Statistical heterogeneity and system heterogeneity exist in both cross-device and cross-silo FL. However, we posit that system heterogeneity is not a major issue in a cross-silo setting, as the organizations are likely to have ample computation/energy resources and stable network connections. 
Instead, cross-silo FL may put more focus on tackling statistical heterogeneity, e.g., training a satisfactory model with highly non-i.i.d. data across different organizations. This is because cross-silo FL applications can have a more stringent requirement in terms of model performance than cross-device scenarios. For example, diagnosis models trained by hospitals are expected to have a sufficiently high accuracy. In cross-device setting, the next-word prediction model can have a relatively lower accuracy requirement since the wrong predictions would cause less harm than improper medical treatment.
Furthermore, among the three types of approaches to tackling statistical heterogeneity, we argue that personalization can be more appropriate than data moderation and clustering for cross-silo FL, due to a lower risk of data leakage.

\subsection{Optimization of Efficiency}
As a distributed machine learning paradigm, federated learning faces the efficiency challenge and it aims to minimize the \textit{computation} and \textit{communication} overheads. 

\subsubsection{Computation Overheads} The local model training process consumes computational resources, e.g., the CPU cycles and energy consumption. To reduce the computation overheads, a plausible method is to select proper and possibly simpler local model structures (e.g., switch from deep neural networks to random forests). This can be tricky, as there is no universal criterion for model selection, and the proper/optimal choice of the model structure is an open problem in machine learning. 

\subsubsection{Communication Overheads} Clients upload their local model updates to the server for aggregation, and then download the updated global model for the next training rounds. This can require a substantial amount of communication resources, e.g., network bandwidths. To reduce the communication overheads, existing studies focus on the following two methods:
\begin{itemize}
	\item \textbf{Model compression}: It compresses the model data by techniques such as quantization, random rotation, and secondary sampling. This can reduce the communication overheads between the server and the FL clients. 
	\item \textbf{Client selection}: It only allows the communication between the server and a selected set of clients. The selection criterion can be based on factors such as historical dropout rates, communication history, and model updates.
\end{itemize}

\textit{Remark}:  The optimization of efficiency is a less major issue in cross-silo FL than in cross-device scenario. First, the cross-silo clients (e.g., companies or organizations) usually have ample computational resources. 
Second, the cross-silo FL clients are expected to use reliable transmission networks (e.g., high-speed wired networks) instead of the mobile cellular networks used by most edge devices in cross-device FL.

\section{Privacy and Security}\label{Privacy_Security}
For cross-silo FL to gain trust and be widely adopted in practice, its privacy and security implications must be well understood and taken care of. In this section, we overview the key existing approaches to privacy protection and security enhancement.

\subsection{Privacy Protection}
Even if federated learning does not directly expose raw data, privacy concerns can arise due to various inference attacks. One type is \textit{membership inference attack}, which aims to check whether certain data points belong to the training set. Canonically, the attacker infers the information via guesswork and training the predictive model to predict the original training data. Another type is \textit{reconstruction attack}, which aims to reconstruct the training data using inference techniques. In particular, reconstruction attack uses GANs to synthesize fake samples that have the same distributions as the training set without accessing the raw data.

To defend against these attacks, many researchers proposed privacy protection strategies \cite{mothukuri2021survey}, which can be summarized into three categories below:

\begin{itemize}
	\item \textbf{Differential privacy}: This is a widely adopted privacy-preserving technique in both academia and industry. The basic idea is to add noise to personal sensitive attributes to protect privacy. In the context of federated learning, the clients add noise to their uploaded model updates to mitigate inference attacks \cite{sun2022profit}. Note that using differential privacy improves clients' privacy protection at the cost of global model performance (e.g., slower convergence and/or accuracy loss). Such a performance degradation may be a major concern in certain cross-silo settings where there is a strict requirement for model performance.
	\item \textbf{Homomorphic encryption}: Such an approach enables certain operations (e.g., addition) to be performed directly on cipher texts without decryption requirements. In cross-silo FL, each client uploads the encrypted local model updates to the server for aggregation, and the result is then sent back to each client for local decryption.  Note that even if homomorphic encryption enhances privacy without hurting the global model performance, it incurs a large amount of computation and communication costs due to the calculation and transmission of the encrypted results.  However, as we mentioned, organizations are likely to have sufficient computation and communication resources. Hence, cross-silo FL applications may prefer homomorphic encryption to differential privacy. 
	\item \textbf{Multi-party computation}: This provides a generic
	approach that enables cross-silo clients to jointly compute
	an arbitrary functionality without revealing their raw data.
    The aggregated model is calculated by exchanging these secret shares among	clients, following well-designed protocols. Similar to homomorphic encryption, multi-party computation involves even more expensive cryptographic operations and hence causes larger communication and computation costs. 
	\end{itemize}

\textit{Remark}: Privacy protection is an even more critical issue in the cross-silo scenario than in the cross-device setting. On the one hand, a wide range of applications of cross-silo FL require proper treatments of human personal data. For example, hospitals can be highly sensitive to their patients' medical records and banks must prevent their customers' financial data from being leaked.  On the other hand, existing cross-silo FL practices usually impose very strict requirements among organizations for data protection. This calls for the further enhancement of privacy in cross-silo FL and deserves sufficient attention. Furthermore, existing approaches cannot achieve a desirable tradeoff between model performance degradation (e.g., differential privacy) and computation/communication overheads (e.g., homomorphic encryption and multi-party computation). It would be important for researchers to develop more sophisticated approaches that can potentially achieve a better tradeoff.

\subsection{Security Enhancement}
Security is another important challenge since federated learning can be vulnerable to adversarial attacks such as poisoning. To be more specific, a proportion of FL clients can be compromised, i.e., either controlled or even owned by some adversary, who acts maliciously to corrupt the training of global model. There are two major types of poisoning attacks. One is \textit{data poisoning}, which aims to fabricate data so that local model updates are wrongly calculated. Another is \textit{model poisoning}, which generates poisoned local updates (e.g., by direct manipulation of model updates) so that the collective model updates deviate from a benign direction. 

To enhance security in FL, numerous studies have proposed various \textbf{robust aggregation} schemes. The key idea is that when doing model aggregation, the central server removes or attenuates the model updates that are judged to be maliciously poisoned based on certain criterion (e.g., the updates that are highly dissimilar to others). See \cite{pillutla2019robust} for an excellent survey on robust aggregation.

\textit{Remark}: While the security issue exists in both cross-device and cross-silo settings,  this can be a less major issue in cross-silo FL. This is because compared to edge devices, organizations are less likely to be compromised by adversaries. In fact, the authors in \cite{shejwalkar2022back} conducted comprehensive experiments and found that poisoning attacks have a  minor impact on cross-silo FL.

\section{Cooperation and Incentives}\label{Cooperation_incentive}

In cross-silo FL, clients are organizations or companies who are likely to have (long-term) development goals and are strategic in maximizing their own benefits. 
It is important to analyze the game theoretical interactions
among cross-silo clients and provide insights on how to promote cooperation.

%
\textit{Remark}: While many of the game-theoretical methods to be mentioned below apply to the cross-device setting, we believe that they are more suitable in cross-silo settings. On one hand, cross-device clients may not have many incentive concerns in certain applications. For example, the federated learning process for next word prediction is embedded in the mobile applications and automatically executed by Google. Hence, there is a small flexibility where mobile users can interact with the server on the incentive issues. On the other hand, cross-silo clients (e.g., organizations) usually have stronger computational resources and reasoning capabilities (than mobile devices in cross-device FL who tend to be short-sighted and boundedly rational), so that it is practical to analyze their interactions based on the full rationality assumption typically assumed in game theoretical analysis. 

\subsection{Data Valuation}
Data valuation (i.e., contribution evaluation) pertains to assessing each client's contribution to cross-silo FL.
 Data valuation was originally intended 
 to explain black-box predictions through the lens of data values. In the context of federated learning, data valuation is used to evaluate the contribution of each participating client, based on which one can design a good incentive mechanism to enhance inter-client cooperation. 

The data valuation methods can be classified into two categories, i.e., quality-agnostic and quality-aware methods. 
\begin{itemize}
	\item \textbf{Quality-agnostic methods}: Quality-agnostic methods measure each client's contribution using metrics such as data sizes and computational resources. For example, \textit{linearly proportional} is a quality-agnostic method that calculates a client's contribution linearly proportional to the data size that the client uses for model training. It requires that the central server knows the clients' used data size (e.g., FedAvg). In case such information is unknown, the server can design auction-based mechanisms to induce clients to truthfully report the information \cite{sarikaya2019motivating}.  Note that quality-agnostic methods are only suitable for the case where clients have i.i.d. data. For the non-i.i.d. case, it is inappropriate to quantify the value of data solely based on data sizes or computational resources. This limits the applicability of quality-agnostic methods.  
	
	\item \textbf{Quality-aware methods}: Quality-aware methods take into account the data quality by retraining models using different combinations of clients' data. Hence, the quality-aware methods are applicable to both the i.i.d. and non-i.i.d. scenarios. There are two types of quality-aware methods. One is \textit{Leave-one-out}, which is a data-counterfactual method that measures a counterfactual of each client' data and examines how much the global model performance changes due to the absence of that data. We will consider the resulting model performance change as the client contribution. 
	Another is \textit{Shapley-value based} methods. 
	It assigns a unique data valuation profile via
	the average marginal model improvement made by each client. More specifically, it compares all the possible training data combinations that include data from one client versus combinations that exclude the data from that particular client. While the Shapley value method has a better empirical performance than leave-one-out, it suffers from exponentially expensive (in the number of participating clients) computation costs. Many empirical methods have been proposed based on Shapley value to reduce the computation costs, such as Monte-Carlo Shapley value and gradient-based Shapley value \cite{ghorbani2019data}.  
	
\end{itemize}
\subsection{Profit Allocation}
Based on the data valuation results, we need to further determine a proper profit/benefit allocation mechanism to incentivize clients to contribute to cross-silo FL. Intuitively, clients who have higher contributions should be allocated more profits to ensure their willingness to cooperate. In the cross-device scenario, the profit allocation mechanism is usually decided by the central server. In cross-silo FL, however, the mechanism can be negotiated by the organizations themselves and enforced by binding agreements and contracts. There are two types of profit allocation mechanisms as follows:

\begin{itemize}
	\item \textbf{Posted-price mechanisms}: A posted-price mechanism specifies how the price (reward) depends on the contribution of each client, and such a mechanism can be announced prior to model training. For example, a posted-price mechanism can offer rewards to a client based on the volume of its data points to be used for model training. 
	\item \textbf{Profit-sharing mechanisms}: A profit-sharing mechanism allocates the total profit/benefit to all the participating clients when they finish the FL task. More specifically, after the cross-silo FL terminates, the clients can utilize a trusted third party to obtain some benefits from the global model (e.g., via selling it in a model trading market). Then, the clients share the profits among themselves using the predefined mechanisms.
\end{itemize}

While existing profit allocation mechanisms mainly focus on enhancing inter-client cooperation (e.g., improving global model performance), they can cause \textbf{fairness issues}, which deserve special attention in cross-silo FL.
There are two facets of fairness.
    \begin{itemize}
	\item \textbf{Inter-client fairness}: One is the fairness among cross-silo clients. For example, different organizations or companies can differ substantially in training-related resources (e.g., data volume and computation capabilities) and market powers. In this case, the ``stronger'' organizations may reap most of the benefits (e.g., due to a larger data volume), discouraging clients to contribute in FL. Some research efforts have been devoted to enhancing fairness. 
	Examples include collaborative fairness methods, which balance the shared profit based on the contribution to the model accuracy, and egalitarian equity methods that shift more focus to client with the worst performance.
	\item \textbf{Model fairness}:  Another facet is on training a fair global model. For instance, hospitals strive to train fair models with medical data collected from geographically varying populations, so that the global model can have a minimum bias toward patients. We believe that it is important to understand how the cross-silo clients' training behaviors affect the fairness of the global model and more research efforts should be devoted along this line. 
	\end{itemize}   
Another important challenge is that even though fairness in FL has been extensively studied, the definitions of fairness substantially vary in literature. Earlier work in \cite{joe2013multiresource} established a unified framework on fair resource allocation, and it is important to borrow their techniques on the unified definition of fairness and derive potential solutions in cross-silo settings.

\subsection{Coalition}
While prior work focuses on the design and optimization of data valuation and profit allocation, it is also important to explore the motivation and incentives regarding how clients form coalitions. In practical cross-silo FL, the organizations may be interested in forming coalitions due to various reasons as follows. 

First, the cross-silo clients usually have heterogeneous and different amounts of data drawn from their own distributions. Consider the case where a group of hospitals collaboratively train a model. Different hospitals are likely to hold heterogeneous data due to, for example, the varying patient populations or variants of the procedure implementation. While the use of federated learning decreases the global model's error due to model variance, the data heterogeneity issue would give rise to a larger model bias. Therefore, some cross-silo clients may have an incentive to form coalitions to jointly construct models (e.g., via sharing local model parameters) such that the eventual global model would be desirable  for clients within the coalition. The authors in \cite{Donahue_Kleinberg_2021} proposed the first game-theoretical framework to analyze how heterogeneous clients optimally form coalitions. While coalition in cross-silo FL receives little attention so far, we believe that more research efforts should be devoted, e.g., to characterize the stability of the coalitions using the concept of core. 

Second, cross-silo clients may have incentives to form coalitions when the market is relatively competitive. Consider that there are numerous banks in the market aiming to provide finance-related services to the potential customers. The banks with limited data may form a coalition to collaboratively train a model. This can help the banks obtain a better model and improve their service quality. As a result, the small banks become more competitive (compared to other bigger banks) and are likely to gain more market power. To the best of our knowledge, there is no prior work along this line and more research attention should be given.

\section{Future Directions and Open Issues}\label{future_work}
The research on cross-silo federated learning is still at its early stage, and there are many interesting open problems to address. We list some of them here.
\subsection{Multi-Objective Optimization}
Current research studies mainly proceed with the perspective of single-objective optimization, e.g., optimizing efficiency/effectiveness, improving privacy/security, or enhancing cooperation, without taking a unified and multi-objective angle. In practice, however, cross-silo FL would encounter multiple challenges at the same time and hence need a more comprehensive approach. For example, differential privacy is a powerful framework to improve  privacy, but it jeopardizes the model performance (i.e., effectiveness) due to noise injection. It would be important to propose a multi-objective optimization framework to understand the interplay among various objectives and derive solutions that give the best tradeoffs.
\subsection{Long-Term Cooperation}
In cross-device FL, clients are mobile edge devices and hence unlikely to form long-term cooperation relations. There are a few reasons. First, edge devices are relatively unstable and they could easily drop out of the training process if the batteries die. Second, the central server which is also the initiator of FL (e.g., Google) usually would dynamically sample part of clients for model training. Third, it is difficult for mobile devices to even identify their cooperating partners due to a vast number of participating clients.

Different from cross-device FL, cross-silo FL provides a hotbed for long-term cooperation. On the one hand, the organizations are relatively stable since they have strong computation and communication resources. On the other hand, the clients themselves initiate the training process, know who their partners are, and hence are likely to stay with the whole training process due to binding agreements/contracts.
A practice example is that WeBank and Swiss Re established a long-term cooperation project in the field of reinsurance. For future work, it would be intriguing to study the strategic behaviors among the cross-silo clients and propose (game-theoretical) mechanisms to promote their long-term cooperation.

\subsection{Business Competition}
Besides cooperation, the cross-silo clients may become business competitors. 
In the domain of finance and insurance, different banks or companies may also strive to induce customers to use their services rather than their competitors.  It would be interesting to study how the business competition problem affects the clients' willingness to cooperate and devote resources for model training. If competition is found to have a hindering impact on the cooperation, then proper mechanisms are needed to mitigate competition and  promote competition,  which ensures the success of cross-silo FL. 


\bibliographystyle{IEEEtran}	
\bibliography{ref,../bib/paper}

\begin{IEEEbiography} [{\includegraphics[width=1in,height=1.25in,clip,keepaspectratio]{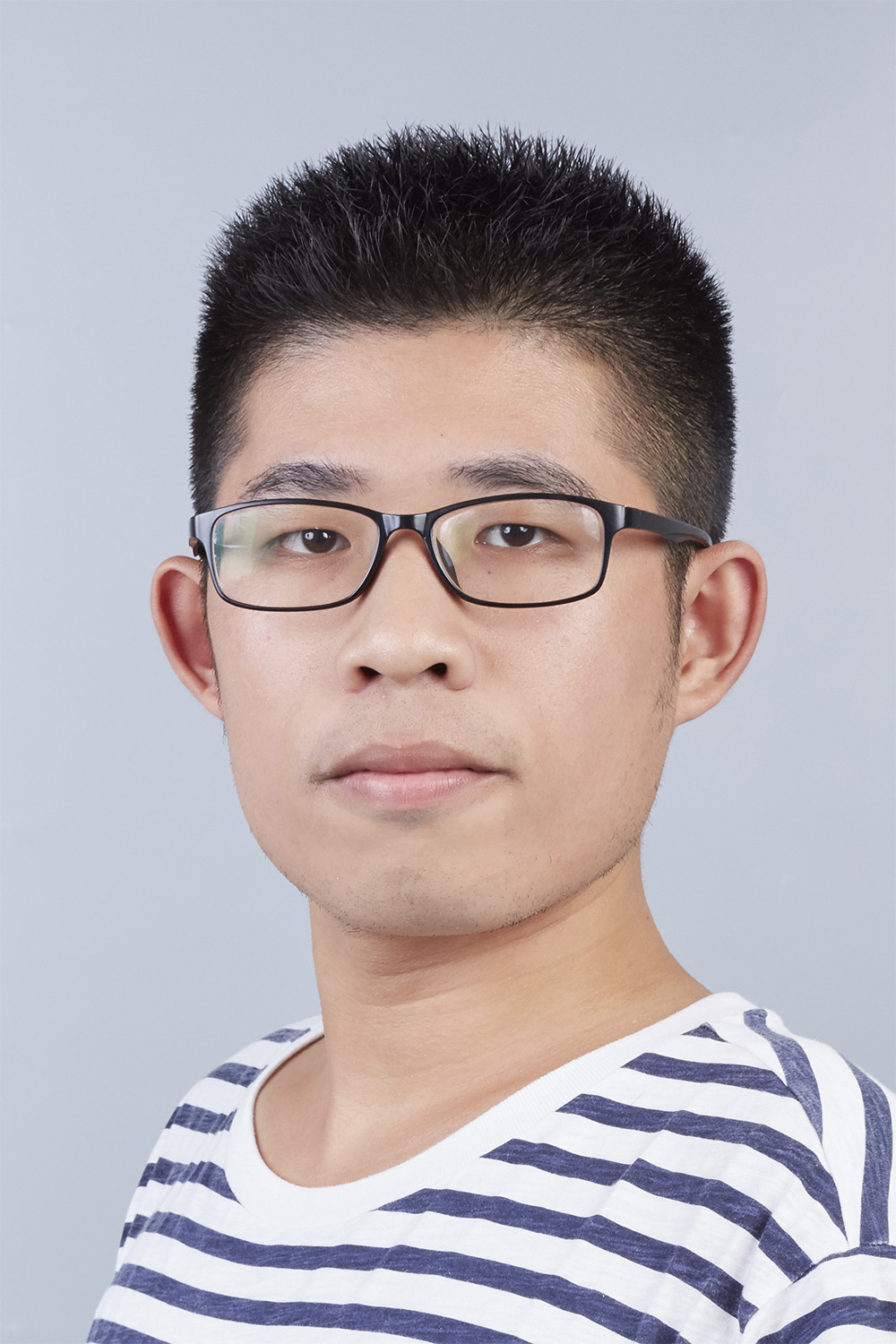}}]
	{Chao Huang} received the Ph.D. degree from the Chinese University of Hong Kong in 2021. During July-November 2021, he was a Post-Doctoral Research Fellow with the Department of Management Sciences, City University of Hong Kong. He is now working as a Post-Doctoral Researcher with the Department of Computer Science, University of California, Davis. His recent research interests span the spectrum of distributed machine learning, network economics, and low-carbon systems.
	
\end{IEEEbiography}

\begin{IEEEbiography} [{\includegraphics[width=1.05in,height=1.6in,clip,keepaspectratio]{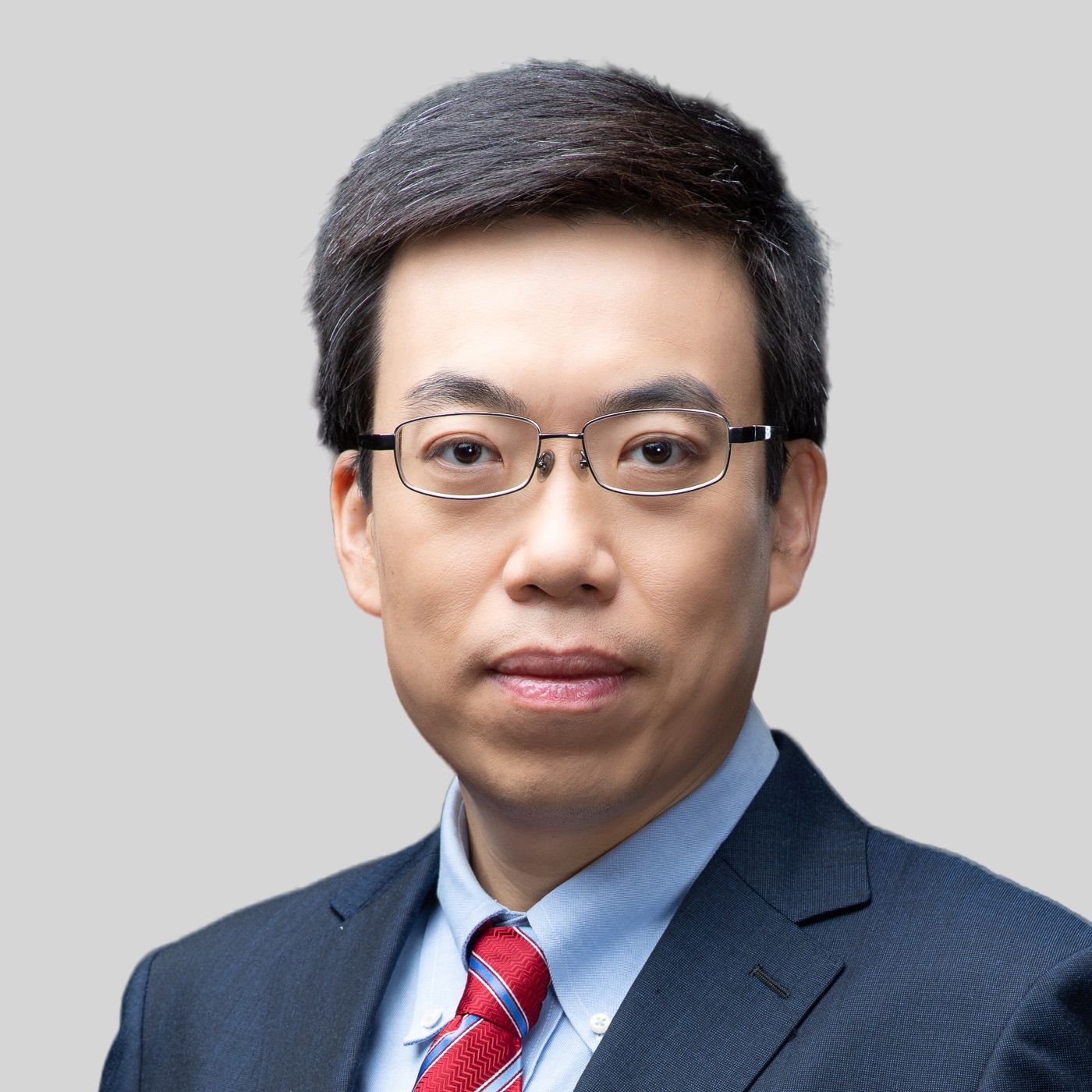}}]
	{Jianwei Huang (F'16)} is a Presidential Chair Professor and Associate Dean of School of Science and Engineering, The Chinese University of Hong Kong, Shenzhen. His research interests are in the area of network optimization, network economics, and network games. He has published more than 300 papers in leading venues, with a Google Scholar citation of 14600+ and an H-index of 61. He has co-authored 10 Best Paper Awards, including the 2011 IEEE Marconi Prize Paper Award in Wireless Communications. He has co-authored seven books, including the textbook on "Wireless Network Pricing." He is an IEEE Fellow, and was an IEEE ComSoc Distinguished Lecturer and a Clarivate Web of Science Highly Cited Researcher. He is the Editor-in-Chief of IEEE Transactions on Network Science and Engineering, and was the Associate Editor-in-Chief of IEEE Open Journal of the Communications Society.
	
\end{IEEEbiography}

\begin{IEEEbiography} [{\includegraphics[width=1in,height=1.4in,clip,keepaspectratio]{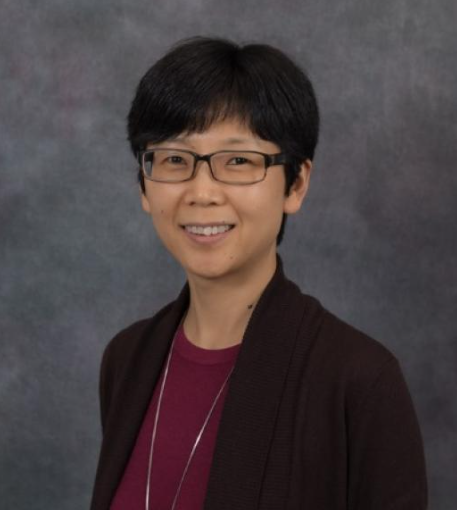}}]
	{Xin Liu} received her Ph.D. degree in electrical engineering from Purdue University in 2002. She is currently a Professor in Computer Science at the University of California, Davis. Her current research interests fall in the general areas of machine learning algorithm development and machine learning applications in human and animal healthcare, food systems, and communication networks. Her research on networking includes cellular networks, cognitive radio networks, wireless sensor networks, network information theory, network security, and IoT systems. 
	
\end{IEEEbiography}

\end{document}